\pdfoutput=1

\documentclass[11pt]{article}
\usepackage{subcaption}
\usepackage{booktabs}
\usepackage{acl}

\usepackage{times}
\usepackage{latexsym}

\usepackage[T1]{fontenc}

\usepackage[utf8]{inputenc}

\usepackage{graphicx}
\usepackage{caption}
\usepackage{listings}
\usepackage{microtype}
\usepackage{xcolor}
\usepackage{inconsolata}
\usepackage{newtxmath} 

\definecolor{codegray}{gray}{0.95}

\lstdefinestyle{mystyle}{
    basicstyle=\ttfamily\small,
    keywordstyle=\color{blue},
    commentstyle=\color{gray},
    stringstyle=\color{red!70!black},
    numbers=none,
    breaklines=true,
    frame=tb,
    captionpos=b
}

\title{JSON Whisperer: Efficient JSON Editing with LLMs}
\author{
 \textbf{Sarel Duanis\textsuperscript{1}} \quad
 \textbf{Asnat Greenstein-Messica\textsuperscript{1}} \quad
 \textbf{Eliya Habba\textsuperscript{1,2}} \\
 \\
 \textsuperscript{1}Lightricks \quad
 \textsuperscript{2}The Hebrew University of Jerusalem \\
[0.5em]
 \texttt{\{sarel, asi, ehabba\}@lightricks.com}
}
\begin{document}
\maketitle



\begin{abstract}
Large language models (LLMs) can modify JSON documents through natural language commands, but current approaches regenerate entire structures for each edit, resulting in computational inefficiency. We present JSON Whisperer, a framework that enables LLMs to generate RFC 6902 diff patches-expressing only the necessary modifications-rather than complete documents.
We identify two key challenges in patch-based editing: (1) LLMs often miss related updates when generating isolated patches, and (2) array manipulations require tracking index shifts across operations, which LLMs handle poorly. To address these issues, we introduce EASE (Explicitly Addressed Sequence Encoding), which transforms arrays into dictionaries with stable keys, eliminating index arithmetic complexities.
Our evaluation shows that patch generation with EASE reduces token usage by 31\% while maintaining edit quality within 5\% of full regeneration with particular gains for complex instructions and list manipulations. The dataset is available at: \noindent\url{https://github.com/emnlp2025/JSON-Whisperer/}

\end{abstract}

\section{Introduction}

Large language models have demonstrated capabilities in generating and manipulating structured data formats, with JSON emerging as a primary interface between natural language commands and programmatic operations. This capability has enabled applications across diverse domains, from web development \cite{voronin2024development} to creative tools \cite{sultan2024visual,kolthoff2025guide}, where users can specify structural modifications through natural language. In film production platforms, for instance, entire projects-comprising scenes, characters, and shot compositions-are represented as JSON documents that creators modify through conversational interfaces.

However, current approaches suffer from a fundamental inefficiency: LLMs regenerate entire JSON structures for even minor edits. This regeneration approach, while straightforward, becomes computationally expensive as document complexity grows. A film project might contain thousands of nested objects representing scenes and shots; changing a single character's name triggers regeneration of the entire structure, consuming substantial computational resources and introducing latency that disrupts creative workflows.

We observe that LLMs can generate RFC 6902 \cite{rfcpatch} diff patches-a standard format for expressing minimal changes to JSON documents-when appropriately prompted. This finding suggests an alternative approach: instead of regenerating complete structures, models could output only the necessary modifications. Yet this seemingly simple solution introduces unexpected challenges.

\begin{figure*}[h]
    \centering
    \begin{subfigure}[b]{0.49\textwidth}
        \centering
        \includegraphics[width=\textwidth]{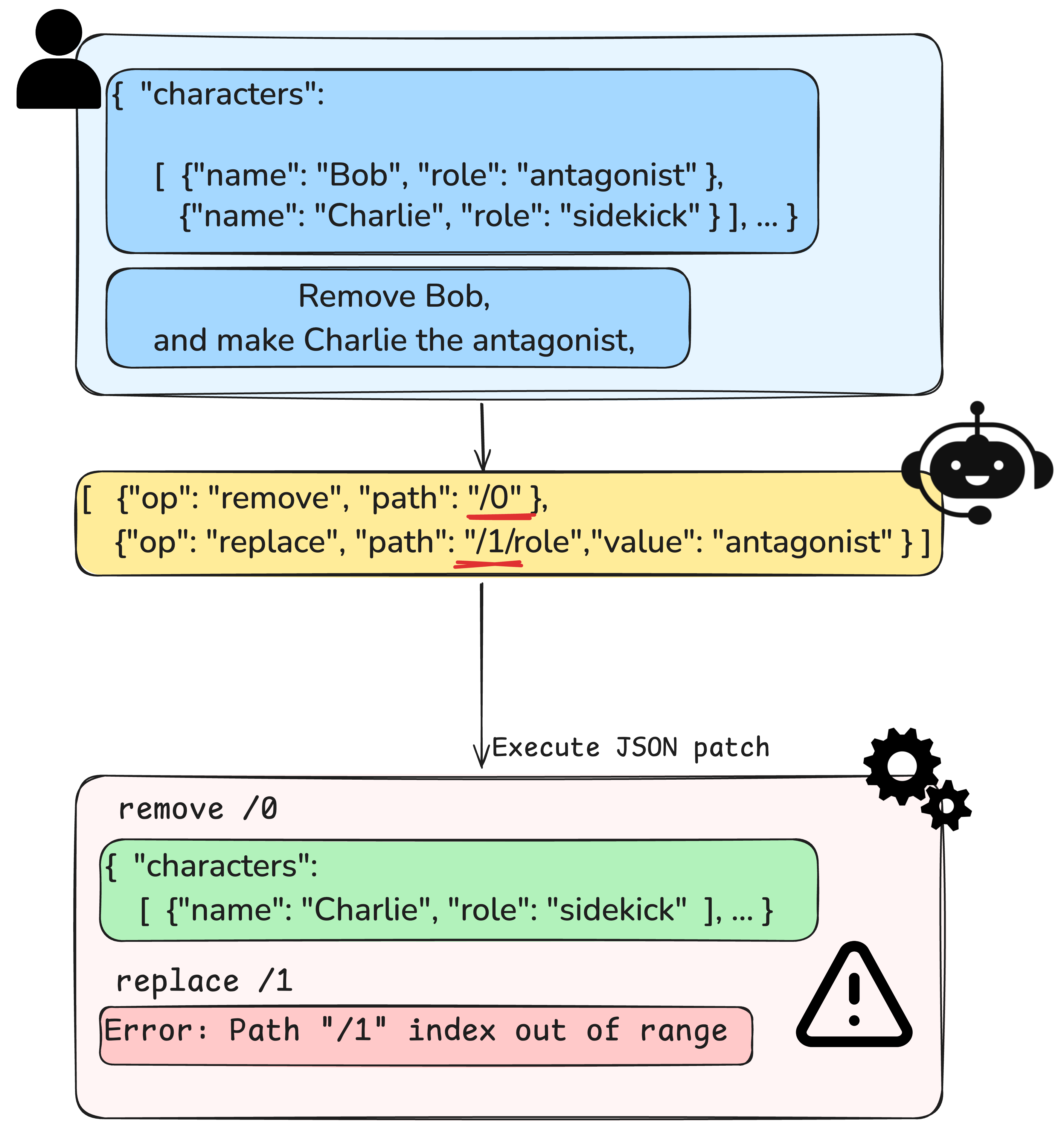}
        \caption{Using normal list indexing}
    \label{fig:sub1}
    \end{subfigure}
    \hfill
    \begin{subfigure}[b]{0.49\textwidth}
        \centering
        \includegraphics[width=\textwidth]{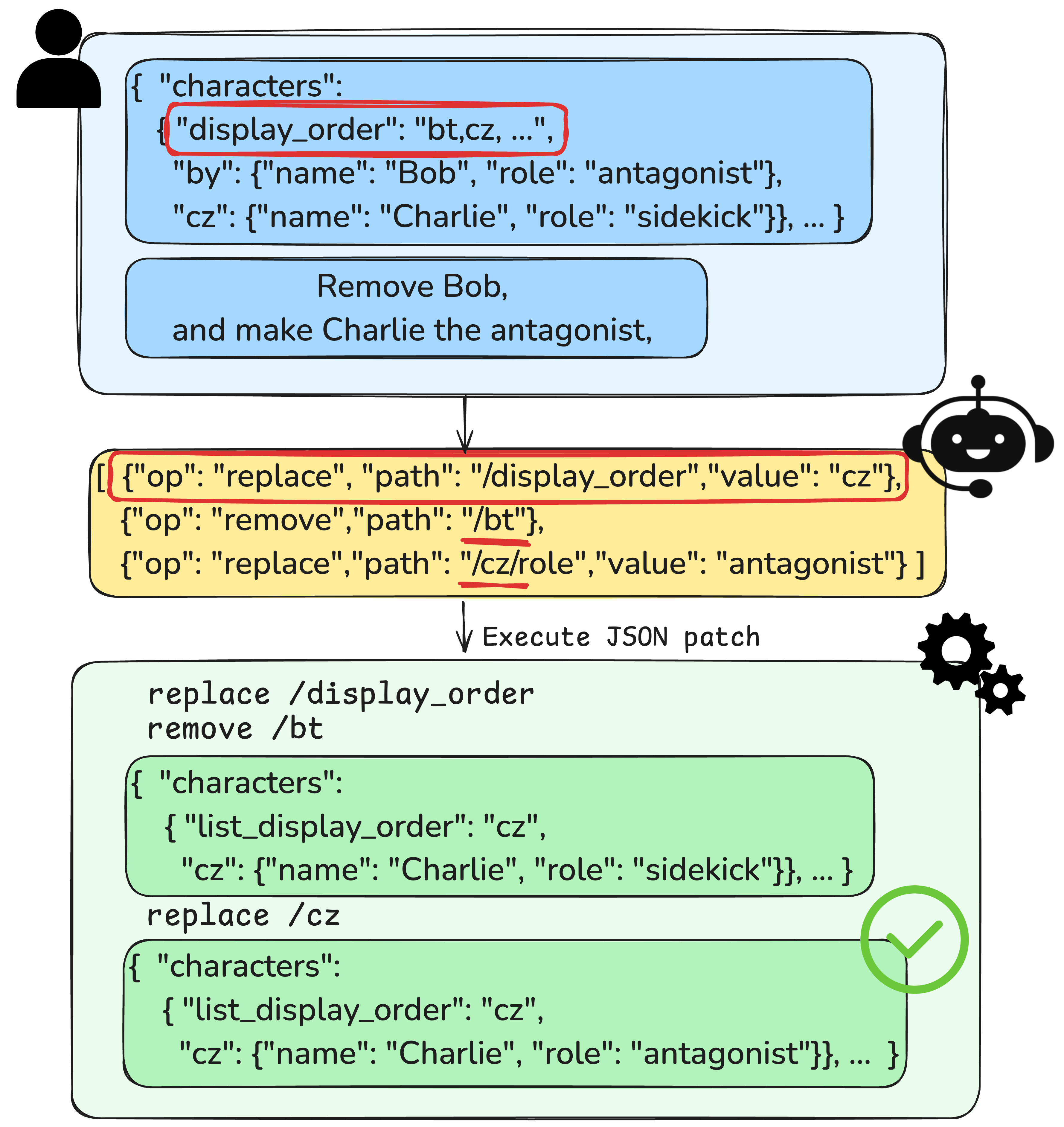}
        \caption{Using EASE}
    \label{fig:sub2}
    \end{subfigure}
    \caption{(\subref{fig:sub1}) Using normal list indexing, LLMs fail to account for index shifts after removing Bob, leading to an error.
(\subref{fig:sub2}) Using EASE, stable keys ensure correct updates, making the patch process execution order-invariant.}
    \label{fig:combined}
\end{figure*}
 
We identify two critical issues that emerge when LLMs generate patches. First, the fragmented nature of patch generation causes models to miss necessary updates. When editing a character's attributes across multiple scenes, models generating patches often update some occurrences while overlooking others-a problem that full regeneration naturally avoids. Second, and more surprisingly, LLMs struggle with array index arithmetic. Consider removing an element from a list: subsequent operations must account for shifted indices as illustrated in Figure~\ref{fig:combined}. Our experiments reveal that even state-of-the-art models frequently miscalculate these shifts or conflate zero-based and one-based indexing.

In this work, we present JSON Whisperer, a framework that enables efficient JSON editing through patch generation while addressing these challenges.

Our contributions are as follows:
\begin{enumerate}
\item we demonstrate that patch-based editing can achieve performance comparable to full regeneration when using appropriate few-shot examples. To this end, we propose a method for synthetically generating such high-quality examples.(Section \ref{sec:framework}).

\item we introduce EASE (Explicitly Addressed Sequence Encoding), a novel representation that eliminates array indexing complexities. EASE transforms positional arrays into key-value mappings with stable identifiers, making all operations order-invariant and removing the need for index arithmetic (Section \ref{sec:ease}).

\item we evaluate our approach on JSON editing scenarios from a film production platform, showing that patch generation with EASE reduces token usage by 31\% while maintaining edit quality within 5\% of full regeneration approaches, with particularly strong improvements for complex instructions and list manipulations (Section \ref{sec:experiments}).
\end{enumerate}
Our work demonstrates that patch-based editing, when combined with EASE encoding, can serve as a practical alternative to full regeneration for JSON manipulation tasks. This combination addresses the fundamental challenge of array indexing that makes standard patch generation error-prone, offering significant efficiency gains for real-world applications where structured data editing is common.

\label{sec:intro}


\section{Background and Definitions}\label{sec:background}
\paragraph{RFC 6902:}
RFC 6902~\cite{rfcpatch} specifies a format for representing modifications to JSON documents as a sequence of operations, as shown in Listing \ref{lst:ops}. Each operation includes:\\
\textbf{op}: The operation type (\texttt{add}, \texttt{remove}, \texttt{replace})\\
\textbf{path}: A JSON Pointer that identifies the target location using forward-slash-separated tokens (e.g., \texttt{/scene/weather}), with zero-based indices for arrays (e.g., \texttt{/users/0/name})\\
\textbf{value}: The data to be used for \texttt{add} and \texttt{replace} operations 

\begin{lstlisting}[caption={This patch will replace the first user name to "John" and add a new user called "Sam"}, label={lst:ops}]
[{"op":"replace", "path":"users/0/name",
  "value": "John"},
 {"op":"add","path":"users/1",
  "value": {"name":"Sam"}}]
\end{lstlisting}


\section{JSON Whisperer Framework}\label{sec:framework}
We present JSON Whisperer, a framework for efficient text-guided JSON editing using LLMs. The framework comprises two main components: diff-based editing using RFC 6902 patches and EASE (Explicitly Addressed Sequence Encoding) for handling ordered lists.

\subsection{Diff-based JSON Editing}
Our method leverages RFC 6902 JSON Patch format to enable targeted modifications. Given an input JSON and a natural language edit request, the LLM generates only the necessary patch operations rather than regenerating the entire structure.

The process follows three steps: (1) the LLM receives the original JSON and edit instruction, (2) it generates an RFC 6902 patch specifying the required changes, and (3) a standard patching algorithm applies these operations to produce the updated JSON.

This approach reduces token usage and latency compared to full regeneration, as the model outputs only the specific modifications needed (e.g., a few patch operations versus an entire JSON document).

\paragraph{Synthetic Dataset Creation.}
To enable effective few-shot learning for diff patch generation, we automatically create a dataset of high-quality examples using a state-of-the-art LLM. Each JSON schema requires its own example set, generated through a fully automated pipeline.
The generation process consists of four steps. First, given a target JSON schema, we prompt Claude-3.5-Sonnet~\cite{claude35sonnet} to generate diverse JSON instances following that schema. Second, for each instance, the LLM creates natural language edit requests simulating realistic user modifications. These requests span from simple field updates to complex structural changes.
Third, for each (JSON, request) pair, we use the same LLM to produce a completely rewritten JSON incorporating all requested modifications. 
Finally, we apply a standard diff algorithm to compute RFC 6902 patches between original and modified JSON pairs. These patches accurately capture the complete transformation specified by each edit request. The resulting dataset provides a pool of examples for few-shot prompting, where each example contains an input (original JSON + edit request) and its corresponding RFC 6902 patch.
Using DSPy~\cite{khattab2024dspy}, we optimize the selection of few-shot examples from this synthetic dataset to maximize evaluation metrics (Section 5.1).

\subsection{EASE - Explicitly Addressed Sequence Encoding}\label{sec:ease}

To improve list manipulation in language models, we introduce EASE - Explicitly Addressed Sequence Encoding, a method for transforming JSON arrays into a more order-independent format. Instead of relying on numerical indices, EASE converts lists into dictionaries where each element is assigned a unique two-character key (e.g., "ax", "cd"). 
The original sequence is preserved using a \emph{display\_order} key (e.g., "ab,cd,ef"); elements can be added, removed, or reordered by replacing only the value of this key.

This approach provides dual order invariance: element storage becomes order-independent through dictionary-based addressing, and operation execution becomes order-independent since patch operations can be applied in any order without affecting the final result.


EASE particularly benefits complex list manipulations where multiple operations would traditionally require careful index arithmetic. By decoupling element identity from order, the encoding allows to focus on the semantic changes rather than bookkeeping details, thereby significantly improving accuracy in edit scenarios.

\begin{lstlisting}[caption={Scene JSON Example}, label={lst:scene}]
{
 "Scene": {
  "voice_over": "Every step brought me closer to the family.",
  "weather": "Sunny",
  "shots": [
   {
    "type": "Wide-shot",
    "action": "Buddy sitting on a sidewalk."
   },
   ...
  ]
 }
}
\end{lstlisting}

\section{Evaluation Setup}\label{sec:experiments}
\paragraph{Models and Prompting Strategies}
We evaluate our JSON editing approach using two models of different scales: GPT-4o-mini (8B parameters) and Claude Sonnet (175B parameters). For each model, we test two prompting strategies: zero-shot prompting without examples and DSPy-optimized few-shot learning using examples from our synthetic dataset. The prompts for standard list indexing and EASE indexing are detailed in Appendix~\ref{sec:Standard List Indexing Model Prompt} and Appendix~\ref{sec:EASE Indexing Model Prompt}, respectively.

\paragraph{Tasks and Dataset}
Our experiments use JSON objects representing complete scenes from an AI-powered platform. Each scene contains shallow fields and a list of shots, where each shot is a nested object, as shown in Listing \ref{lst:scene}.

We generate a dataset of approximately 400 examples using Claude-Sonnet, covering diverse JSON editing scenarios across four complexity levels:

\begin{itemize}
  \item \textbf{Simple}: Affects a single field without impacting others.\\
    \textit{e.g., "Change the weather to 'Partly cloudy with a light breeze.'"}
  
  \item \textbf{Creative}: Involves generating new content that doesn't exist in the original JSON.\\
    \textit{e.g., "Add one more shot to the scene."}

  \item \textbf{Complex}: Requires modifying multiple related elements while preserving logical consistency.\\
    \textit{e.g., "Change the scene so that Melody's ex-partner is in the audience, and she notices him halfway through her performance."}
  
  \item \textbf{List Manipulation}: Involves ordering, filtering, or updating items within a list.\\
    \textit{e.g., "Remove every third shot from the scene."}
\end{itemize}

\paragraph{Evaluation Metrics}
We assess model performance using three key metrics. First, we measure structural accuracy through the F1 score of patch operations, comparing only the 'op' and 'path' fields against dataset labels while excluding 'value' fields due to their potential for multiple valid variations. Second, we evaluate the semantic quality of generated values using an LLM-as-a- judge,(see Appendix~\ref{app:judge_prompt} for the prompt used). Finally, we track the patch execution success rate as the proportion of patches that do not fail to apply due to syntax errors, missing fields, or invalid paths.

\section{Results}
\begin{figure}[h]
    \raggedright
   \begin{minipage}{0.48\textwidth}
        \centering
        \label{fig:diff_patch_results}
    \includegraphics[width=\linewidth]{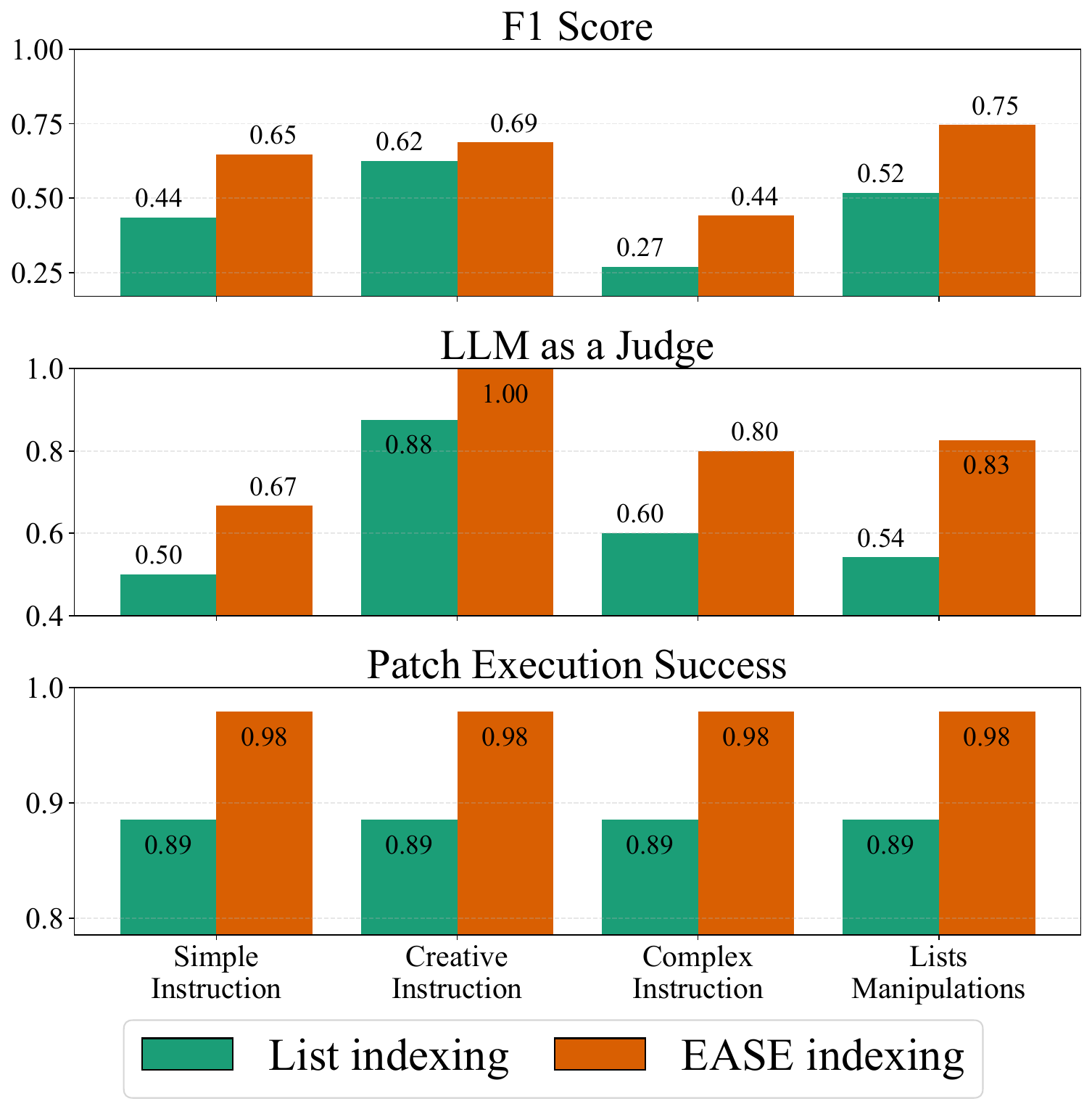}
    \end{minipage}
   \caption{EASE Encoding Outperforms Standard List Indexing, broken down by different request types generated by GPT-4o-Mini}
   \label{fig:ease_vs_lists}
\end{figure}

\paragraph{EASE Encoding Outperforms Standard List Indexing}
EASE encoding consistently outperforms standard list indexing across all instruction categories, with the most pronounced improvements in complex instructions and list manipulations. Figure~\ref{fig:ease_vs_lists} breaks down EASE encoding contributions across different request types. The comparison shows EASE improved accuracy and reduced errors across all instruction categories, while standard list indexing produces suboptimal results due to index tracking challenges. This approach particularly benefits scenarios involving multiple list operations that would traditionally require careful index arithmetic, leading to improved f1 score, semantic accuracy and patch execution.

\begin{figure}[h]
    \raggedright

    \begin{minipage}{0.48\textwidth}
        \centering
    \includegraphics[width=\linewidth]{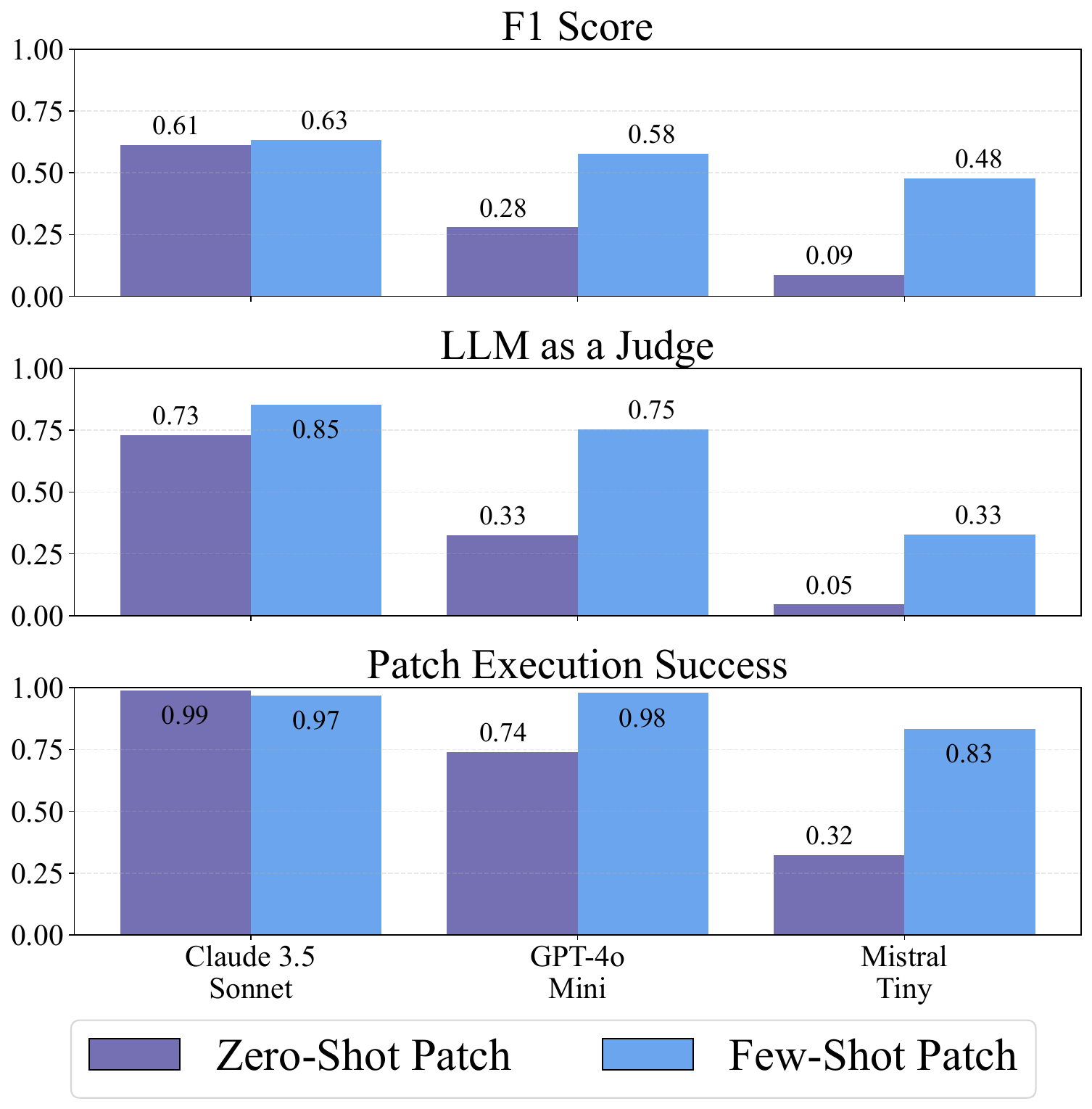}
    \end{minipage}
   
   \caption{Using Synthesized Few-shots provide substantial performance improvements across models}
   \label{fig:zero_vs_few_shots}
\end{figure}

\paragraph{Few-Shot Learning Enables Effective Patch Generation}
Synthetic few-shot examples provide substantial performance improvements in JSON patching tasks, having the most significant impact on performance. As demonstrated in Figure~\ref{fig:zero_vs_few_shots}, comparing zero-shot performance with synthetic few-shot learning using standard list indexing reveals significant accuracy gains across model sizes. This improvement demonstrates that LLMs can effectively learn patch generation patterns from automatically generated examples, even when using conventional list indexing approaches.

\begin{figure}[h]
    \centering
    \begin{minipage}{0.48\textwidth}
        \centering
        \includegraphics[width=\linewidth]{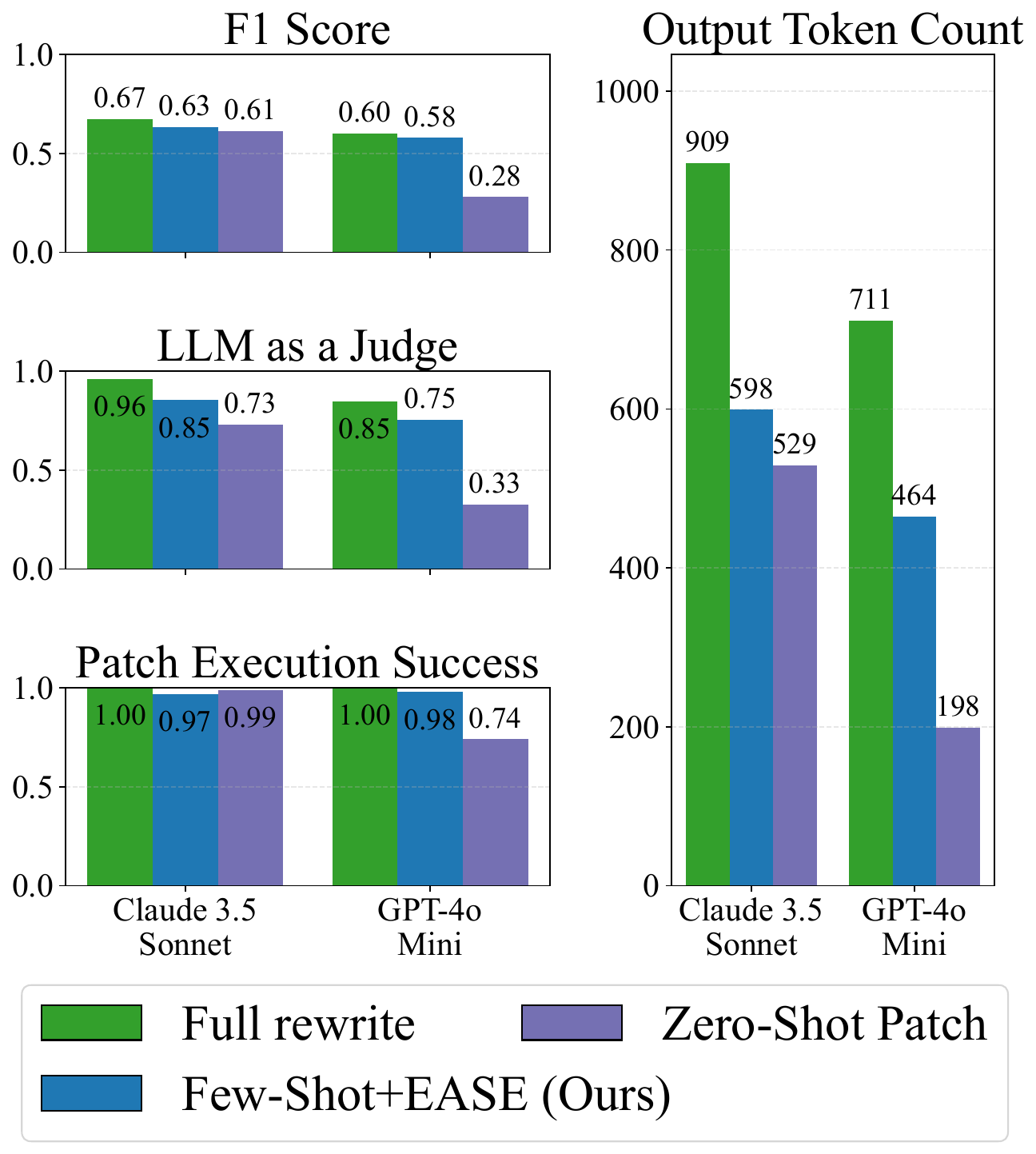}
    \end{minipage}%
    \caption{Our method achieves comparable performance within a 5\% margin to full regeneration while reducing token usage by 31\%}
\label{fig:full_json_results}
\end{figure}

\paragraph{Patch Generation Matches Full Regeneration Performance}
Our diff patching approach with synthesized few-shot examples and EASE achieves comparable performance within a 5\% margin to full regeneration while reducing token usage by 31\%, as shown in Figure~\ref{fig:full_json_results}. The approach demonstrates effectiveness across both model sizes, with smaller models achieving performance comparable to larger ones when using our method (prompts for full regeneration are provided in Appendix~\ref{sec:Full JSON Regeneration Model Prompt}).

\paragraph{Substantial Cost and Efficiency Gains}
The efficiency gains are twofold, as shown in Table~\ref{tab:model-comparison}: first, by enabling smaller, more cost-effective models to perform comparably to larger ones, and second, by reducing token consumption for each operation. These benefits compound to create substantial cost and time savings in production environments. The token reduction translates directly to faster processing times and lower computational costs, making the approach practical for real-world applications where efficiency is paramount.

\begin{table}[h!]
  \centering
  \caption{Costs \& Time Comparison}
  \label{tab:model-comparison}
  \footnotesize
  \begin{tabular}{lccc}
    \toprule
    & Claude & 4o-mini  \\
    \midrule
    \multicolumn{4}{l}{\textbf{Token Count (In/Out)}} \\
    Full & 918 / 918 & 677 / 677 \\
    Diff & 1,890 / 529 & 1,500 / 464 \\
    \midrule
    \multicolumn{4}{l}{\textbf{Time (seconds)}} \\
    Full & 11.62 & 8.26   \\
    Diff & \textbf{6.70} & \textbf{5.66} &   \\
    Improvement & \textbf{42.3\%} & \textbf{31.5\%} \\
    \midrule
    \multicolumn{4}{l}{\textbf{Cost (\$ / K requests)}} \\
    Full & 13.8\$ & 0.41\$ \\
    Diff & \textbf{7.94\$} & \textbf{0.28\$} & \\
    Improvement & \textbf{42.5\%} & \textbf{31.7\%} \\
    \bottomrule
  \end{tabular}
  \begin{flushleft}
    \footnotesize{Model pricing: Claude (\$15.00 per M Out tokens), GPT-4o-mini (\$0.60 per M Out tokens).\\
    TPOT (Claude) = 0.0127, TPOT (GPT-4o-mini) = 0.0122. Data source: \url{https://artificialanalysis.ai/}}
  \end{flushleft}
\end{table}

\section{Related Work} \label{sec:related}

\subsection*{LLMs for Structured Data Manipulation}
In recent years, significant research has been directed toward guiding large language models (LLMs) to generate structured outputs, such as JSON~\cite{geng2025generating,shorten2024structuredrag}. Recent studies further show promising results in LLM-guided generation of JSON objects that represent visual entities such as websites, virtual reality environments, or video effects~\cite{chen2025llmer,voronin2024development,kolthoff2025guide,sultan2024visual}. 

Schema-guided decoding has emerged as a central technique for ensuring that model outputs conform to well-defined formats. Early approaches such as PICARD~\cite{scholak2021picard} demonstrated how constrained autoregressive decoding can enforce syntactic validity by rejecting illegal continuations. These advances highlight the importance of structural validity as a guiding principle in structured text generation. 

Complementing these efforts, research on streaming JSON parsers has focused on reducing the computational overhead of handling large structured documents. Notable examples include Mison~\cite{li2017mison}, which introduced speculative, field-aware parsing to bypass irrelevant fields, and subsequent extensions that further improved throughput using SIMD and incremental indexing~\cite{langdale2019simdjson}. Unlike these studies, the approach presented here enables efficient manipulation of structured data objects such as JSON by patch generation~\cite{jsondiffpatch}, thereby reducing both cost and latency.

\subsection*{Automated Patch Generation}
Research on automated patch generation and program repair by LLM-guided code updates and bug fixes provides valuable insights for JSON editing with LLMs. Fan et al.~\cite{fan2024exploring} explored LLMs’ capabilities for code-change-related tasks, showing that few-shot learning and fine-tuning approaches are promising for such scenarios. Zhang et al.~\cite{zhang2025patch} demonstrate how LLMs can generate patches for code, which closely parallels the concept of creating JSON diff patches for structured data. By leveraging similar methodologies, LLMs can be guided to generate targeted modifications within JSON structures rather than regenerating entire objects, thereby improving both efficiency and accuracy in structured data editing.

\section{Conclusion}
This paper introduces an efficient approach for text-guided JSON editing using LLMs to generate only the necessary modifications instead of regenerating entire JSON objects. The method leverages the RFC 6902 JSON patch standard, significantly reducing token usage, computation costs, and latency, which are key factors for scalable AI-driven applications.
To improve accuracy, a framework for synthetic data generation was developed alongside EASE, a novel encoding method that enhances LLMs' ability to handle ordered lists in JSON structures. Experimental results demonstrate that the approach reduces token generation by over 30\%, while maintaining JSON edit quality within a 5\% margin of full JSON regeneration.

For future work, the method could be extended to YAML and Git diff formats, while fine-tuning open-source LLMs might improve accuracy. Given the widespread use of JSON in software development, this research contributes to more efficient and intelligent AI-powered editing tools for structured data manipulation.

\section{Limitations}
Our approach faces two key limitations. First, both training and evaluation rely on synthetic data, which may not fully reflect real-world JSON editing scenarios. This limitation could be mitigated once user data consented for training is available. Second, while our method reduces output token usage and latency, the in-context learning approach increases input token overhead as JSON objects grow larger. This increase could offset the cost savings and ultimately lead to context window limitations. These challenges could be addressed through model fine-tuning, which would eliminate the need for few-shot examples and enable more efficient processing of larger JSON documents.

\clearpage
\bibliography{custom}
\clearpage
\appendix
\onecolumn
\section{Prompts Used in Experiments}
\label{sec:appendix}
This appendix contains the complete prompts used for the three different approaches evaluated in our experiments: Standard List Indexing, EASE Indexing, and Full JSON Regeneration.

\subsection{Standard List Indexing Model Prompt}
\label{sec:Standard List Indexing Model Prompt}

\begin{small}
\begin{verbatim}
You are a helpful movie production assistant, integrated inside a generative AI video app.
You are a master of your craft, and confident with your choices. You know perfectly well 
how to edit a scene and cast characters.

Given an input json representing an object in a short film and a user command,
your job is to return the json-patches needed to be made on the input json in order 
to edit it fully and hermetically according to the command.
Try to understand the users desired result
Make as many updates as possible, in order to fully and hermetically follow the 
edit command.

Your Jsons need to be sensitive in terms where it is not changing things that 
shouldn't be changed, and changing things that need to be changed, including their 
implications, for example:
- If the director commands you to add a shot to the scene - Be creative enough to 
  make a shot with a description that suits the scene, while not changing the rest 
  of the shots in the scene.
- If the director commands you to change scene's location - Change the location and 
  anything implied from this change (e.g, if the location has changed from a beach 
  to a jungle, and shot 3 has waves SFX, change it to Wind).
- If the director commands you to change shot 3 and shot 2 to be more funny, don't 
  just add keywords such as laughter and funny, but be creative enough to make this 
  shot funnier in the context of the scene.

If the command of the user is unrelated to the json and your task ignore it and 
return is_unsupported action flag as True
If the updated json schema or any sub json schema is different from the input json 
or sub json schema, is_unsupported flag is True, otherwise False

Output Fields:
- rationale: one line string start with 'let's think step by step, we need to...' 
  explain the problem and how you are going to solve it
- json_diff_patch: json-patches (RFC 6902) needed to be made on the input json in 
  order for the json to match the command, all lists indices are always 0 based.
- is_unsupported: True if the command is unrelated to the json or the updated json 
  schema is different from the input json, otherwise False
\end{verbatim}
\end{small}

\subsection{EASE Indexing Model Prompt}
\label{sec:EASE Indexing Model Prompt}
\begin{small}
\begin{verbatim}
You are a helpful movie production assistant, integrated inside a generative AI video app.
You are a master of your craft, and confident with your choices. You know perfectly well 
how to edit a scene and cast characters.

Given an input json representing an object in a short film and a user command,
your job is to return the json-patches in RFC 6902 needed to be made on the input json 
in order to edit it fully and hermetically according to the command.
Try to understand the users desired result
Make as many updates as possible, in order to fully and hermetically follow the 
edit command.

A sub json contains a "list_display_order" is an ordered dict,
list_display_order determines the item order, not the dictionary order.
Keys are random two-letter strings (e.g., xy), with no inherent order.
Every item is accessed by its key, and the order is determined by the list_display_order.
To add, remove or move items, you must update list_display_order as a single string,
for example:
given a dictionary:
 list_display_order: "xk,xy,np,cv"
moving the second item to the first place, and the last item to the second place 
will result:
 list_display_order: "xy,cv,xk,np"

Adding a new item without any specific ordering will add the item as the last, 
result updating the list_display_order as:
list_display_order: "xy,cv,xk,np,rt"
and adding the item with random key "rt" to the dictionary.

Adding a new item in to the third place will result updating the list_display_order as:
 list_display_order: "xy,cv,yi,xk,np,rt"
 and adding the item with random key "yi" to the dictionary.

Removing the first item will result updating the list_display_order as:
 list_display_order: "cv, rt, xk, np"
 and removing the item with key "xy" from the dictionary.
In any case existing mapping between keys and values should be kept.

Your Jsons need to be sensitive in terms where it is not changing things that 
shouldn't be changed, and changing things that need to be changed, including their 
implications, for example:
- If the director commands you to add a shot to the scene - Be creative enough to 
  make a shot with a description that suits the scene, while not changing the rest 
  of the shots in the scene.
- If the director commands you to change scene's location - Change the location and 
  anything implied from this change (e.g, if the location has changed from a beach 
  to a jungle, and shot 3 has waves SFX, change it to Wind).
- If the director commands you to change shot 3 and shot 2 to be more funny, don't 
  just add keywords such as laughter and funny, but be creative enough to make this 
  shot funnier in the context of the scene.

If the command of the user is unrelated to the json and your task ignore it and 
return is_unsupported action flag as True
If the updated json schema or any sub json schema is different from the input json 
or sub json schema, is_unsupported flag is True, otherwise False

Output Fields:
- rationale: one line string start with 'let's think step by step, we need to...' 
  explain the problem and how you are going to solve it
- json_diff_patch: json-patches (RFC 6902) needed to be made on the input json in 
  order for the json to match the command, all lists indices are always 0 based.
- is_unsupported: True if the command is unrelated to the json or the updated json 
  schema is different from the input json, otherwise False
\end{verbatim}
\end{small}

\subsection{Full JSON Regeneration Model Prompt}
\label{sec:Full JSON Regeneration Model Prompt}

\begin{small}
\begin{verbatim}
You are a professional movie director,
You are creative, a master of your craft, and confident with your choices.
You know perfectly well how to edit a scene and cast characters.
You get a json representing an object in a movie editing software and a user command
and returns a label containing the updated json according to the command,
you should cover all changes that has to be done in the json object,
write the updated json as a professional director,
make as many updates as possible, in order to fully and hermetically follow the 
edit command.

Your Jsons need to be sensitive in terms where it is not changing things that 
shouldn't be changed, and changing things that need to be changed, including their 
implications, for example:
- If the director commands you to add a shot to the scene - Be creative enough to 
  make a shot with a description that suits the scene, while not changing the rest 
  of the shots in the scene.
- If the director commands you to change scene's location - Change the location and 
  anything implied from this change (e.g, if the location has changed from a beach 
  to a jungle, and shot 3 has waves SFX, change it to Wind).
- If the director commands you to change shot 3 and shot 2 to be more funny, don't 
  just add keywords such as laughter and funny, but be creative enough to make this 
  shot funnier in the context of the scene.

If the command of the user is unrelated to the json and your task ignore it and 
return is_unsupported action flag as True
If the updated json schema or any sub json schema is different from the input json 
or sub json schema, is_unsupported flag is True, otherwise False
Return the updated json according to the command without comments in it

Output Fields:
- rationale: one line string start with 'let's think step by step, we need to...' 
  explain the problem and how you are going to solve it
- updated_json: the updated json according to the command keeping the same schema 
  as the input json
- is_unsupported: True if the command is unrelated to the json or the updated json 
  schema is different from the input json, otherwise False
\end{verbatim}
\end{small}

\subsection{LLM-as-a-judge Prompt}\label{app:judge_prompt}
This prompt is used to evaluate which of two JSON editing approaches produces 
higher quality results for a given editing command.
\begin{small}
\begin{verbatim}
Compare the quality of two updated jsons given the input json and command. 
along the specified dimension.

Input Fields:
- original_json: original json
- w_json: w json (first comparison candidate)
- v_json: v json (second comparison candidate)  
- user_command: the command to apply on the json
- quality_question: The question should be answered assessing the quality of the 
  json json-patches given the command, and the input json
  
quality_question:
"which json is of a higher quality update according to the command in term of 
the written content?"

Output Fields:
- quality_answer: w / v / tie (only)

\end{verbatim}
\end{small}

\end{document}